\documentclass[final]{cvpr}

\usepackage{times}
\usepackage{epsfig}
\usepackage{graphicx}
\usepackage{amsmath}
\usepackage{amssymb}

\usepackage{pifont}
\usepackage[table]{xcolor}
\usepackage{colortbl}
\usepackage{enumitem}
\usepackage{caption}
\usepackage{comment}
\usepackage{adjustbox}
\usepackage{makecell}
\usepackage[ruled]{algorithm}
\usepackage{algpseudocode}
\usepackage{booktabs}
\usepackage{tabu}

\newcommand{\ieno}{\textit{i}.\textit{e}.}
\newcommand{\egno}{\textit{e}.\textit{g}.} %there is no space
\definecolor{Gray}{gray}{0.95}
\newcolumntype{a}{>{\columncolor{Gray}}c}
\newcolumntype{b}{>{\columncolor{white}}c}

\makeatletter
\def\thickhline{%
	\noalign{\ifnum0=`}\fi\hrule \@height \thickarrayrulewidth \futurelet
	\reserved@a\@xthickhline}
\def\@xthickhline{\ifx\reserved@a\thickhline
	\vskip\doublerulesep
	\vskip-\thickarrayrulewidth
	\fi
	\ifnum0=`{\fi}}
\makeatother

\newlength{\thickarrayrulewidth}
\usepackage[pagebackref=true,breaklinks=true,colorlinks,bookmarks=false]{hyperref}

\def\cvprPaperID{3555} % *** Enter the CVPR Paper ID here
\def\confYear{CVPR 2021}

\begin{document}
	
\title{MetaAlign: Coordinating Domain Alignment and Classification for Unsupervised Domain Adaptation}

\author{
{Guoqiang Wei{$^{1}$}\thanks{This work was done when Guoqiang was an intern at MSRA.}} \qquad Cuiling Lan{$^{2}$}\thanks{Corresponding Author.} \qquad   Wenjun Zeng{$^{2}$} \qquad  Zhibo Chen{$^{1\dagger}$} \qquad\\
\normalsize
$^{1}$\	University of Science and Technology of China ~~ $^{2}$\,Microsoft Research Asia\\
\normalsize
{\tt\small wgq7441@mail.ustc.edu.cn\quad \{culan,wezeng\}@microsoft.com\quad chenzhibo@ustc.edu.cn}
}
\maketitle

\begin{abstract}
	For unsupervised domain adaptation (UDA), to alleviate the effect of domain shift, many approaches align the source and target domains in the feature space by adversarial learning or by explicitly aligning their statistics. However, the optimization objective of such domain alignment is generally not coordinated with that of the object classification task itself such that their descent directions for optimization may be inconsistent. This will reduce the effectiveness of domain alignment in improving the performance of UDA.
	In this paper, we aim to study and alleviate the optimization inconsistency problem between the domain alignment and classification tasks. 
	We address this by proposing an effective meta-optimization based strategy dubbed MetaAlign, where we treat the domain alignment objective and the classification objective as the meta-train and meta-test tasks in a meta-learning scheme. MetaAlign encourages both tasks to be optimized in a coordinated way, which maximizes the inner product of the gradients of the two tasks during training. Experimental results demonstrate the effectiveness of our proposed method on top of various alignment-based baseline approaches, for tasks of object classification and object detection. MetaAlign helps achieve the state-of-the-art performance.
\end{abstract}

\section{Introduction}

With the advance of deep convolutional neural networks (CNN), computer vision tasks such as image classification and object detection have gained significant improvement \cite{krizhevsky2012imagenet, ren2015fasterrcnn, he2017maskrcnn}. In general, the trained models perform well on the testing dataset of which the distribution bears a resemblance to that of training dataset.
However, in many practical scenarios, directly applying such trained models to a new domain usually suffers from significant performance degradation. There exist differences in data characteristics/distributions between the training and testing domains, which are known as \textit{domain shift}~\cite{torralba2011unbiased, sun2016return}. This makes it hard to directly transfer knowledge learned from source to target. Annotation on the samples of the target domain can alleviate this problem but is expensive and time-consuming. Without the requirement of annotation on target samples, unsupervised domain adaption (UDA) attracts a lot of attention, which allows us to learn to adapt the model trained on source to target by exploiting the unlabeled target samples.  

\begin{figure*}[t]
	\centering
	\includegraphics[width=0.99\textwidth]{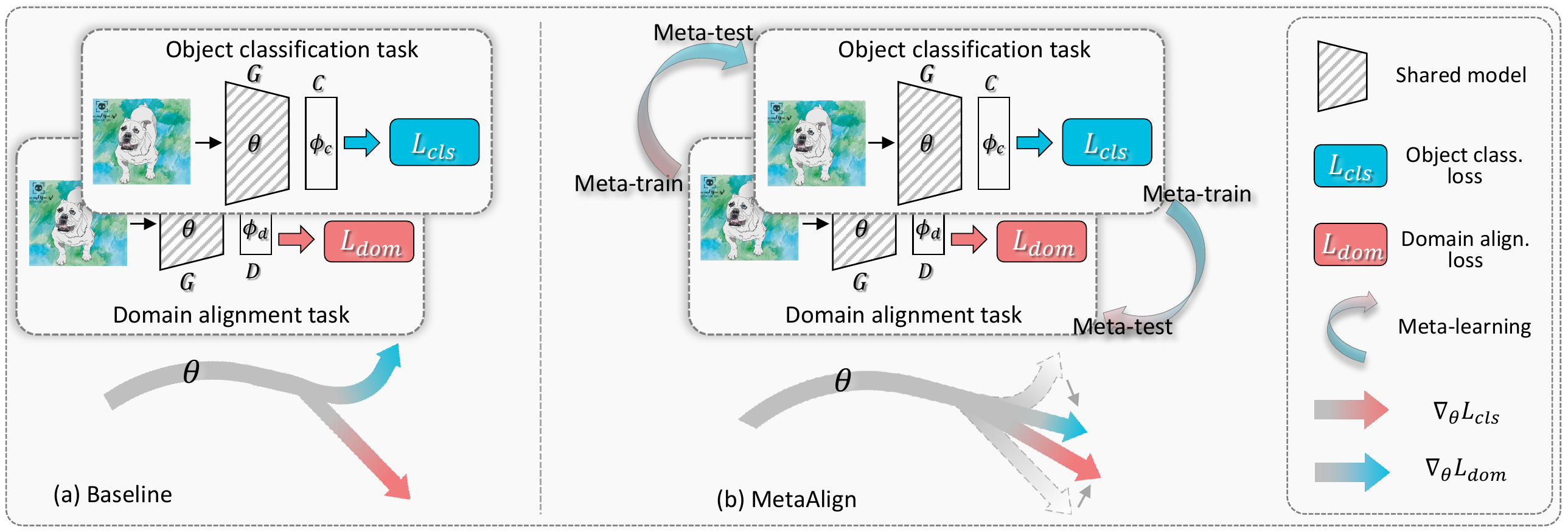}
	\vspace{-0.3cm}
	\caption{Illustration of our MetaAlign strategy which aims to encourage the optimization consistency between the domain alignment task and the object classification task for efficient UDA. 
		(a) Previous approaches directly combine the optimization objective functions of the two tasks together (\ieno, $\mathcal{L}_{dom}+\mathcal{L}_{cls}$), where the descent directions for optimizing the shared network parameters $\theta$ from the two tasks may be inconsistent. (b) In contrast, we treat one of these two tasks as meta-train task and the other as meta-test task. We leverage this meta-optimization based strategy to enforce the consistency between their optimization gradients w.r.t. $\theta$.
		MetaAlign is generic and applicable to various domain alignment based UDAs. }
	\label{fig:pipeline}
	\vspace{-0.3cm}
\end{figure*}

There has been a large spectrum of UDA methods developed in the literature. The major line among them attempts to align the distributions of source and target domains by learning domain-invariant representations, through directly minimizing the discrepancy between feature distributions of two domains \cite{tzeng2014deep,long2015learning,sun2016deep,sun2016return,geng2011daml,zhang2019domainsymnet} or adversarially learning to enforce the feature representations to be indistinguishable by a domain discriminator \cite{ganin2015dann, ganin2016domaindann,pei2018multi,long2018cdane,wang2019transferabletada,chen2020adversarialadla,cui2020gradually, matsuura2020domainmmld}. 
The former category of methods usually align domain distributions by employing explicit distribution similarity metrics, \textit{e.g.}, momentum distance \cite{peng2019momentm3sda, zellinger2017central}, or the second-order correlation \cite{sun2016return,sun2016deep,peng2018synthetic}, between the source and target domains. The latter one borrows ideas from Generative Adversarial Networks \cite{goodfellow2014generative} and use adversarial training to learn aligned feature representations. However, these alignment constraints/strategies are actually not designed specially for the object classification task. There is a lack of efficient coordination between the optimizations of these two tasks. 
During training, the optimization 
procedure of alignment may be inconsistent with that of the object classification task itself, which could hurt the learning of discriminative object features for classification and thus results in inferior object classification performance.

In this work, we aim to address this pervasive problem/challenge faced by alignment-based unsupervised domain adaptation methods, \ieno, the optimization inconsistency between the domain alignment task and the classification task itself. We propose a meta-learning \cite{schmidhuber1995learning, thrun2012learning} based method dubbed MetaAlign to mitigate such inconsistency. Particularly, as illustrated in Fig.~\ref{fig:pipeline}~(b), we treat the domain alignment objective and the classification objective as two tasks in a meta-learning scheme, where we take one task as meta-train task to optimize the network and meanwhile we validate 
the optimization result on the other task (\ieno, meta-test task) for the same set of samples, during training.
Meta-optimization across these two tasks encourages both to be optimized in a coordinated way. 
The theoretical analysis reveals that MetaAlign achieves this optimization coordination by maximizing the inner product of the gradients of the two tasks during training. 

We summarize our contributions as follows:
\begin{itemize}[leftmargin=*,noitemsep,nolistsep]
	\item We pinpoint the problem/challenge in existing alignment-based UDA methods: the optimization inconsistency between the domain alignment task and the classification task itself. To address the problem, we propose a meta-optimization based strategy named MetaAlign to mitigate the inconsistency.   
	
	\item The proposed MetaAlign strategy is generic and can be applied to \emph{various domain alignment based UDA methods} for \emph{object classification and detection} to enforce domain alignment while preserving the discrimination power of features for the recognition task. 
\end{itemize}

We validate the effectiveness of MetaAlign on the image classification (unsupervised domain adaptation and domain generalization) and object detection (unsupervised domain adaptation) tasks. For image classification, we implement MetaAlign on top of various domain alignment based UDA baselines. Extensive experimental results demonstrate the effectiveness and applicability of MetaAlign and we achieve the state-of-the-art performance.

\section{Related Work}
\label{sec:related}

\noindent\textbf{Unsupervised Domain Adaptation.} 
Unsupervised Domain Adaptation (UDA) aims to transfer the knowledge from a labeled source domain to an unlabeled target domain. Abundant UDA works focus on object classification or use it for their investigations. 
The mainstream approaches tend to address UDA by learning domain-invariant representation, to which our proposed method belongs. These approaches can be categorized into two categories.

One category explicitly reduces the domain discrepancy measured by some distribution discrepancy metrics. \cite{tzeng2014deep,long2015learning, long2017deep, yan2017mind} measure the domain similarity in terms of Maximum Mean Discrepancy (MMD) \cite{borgwardt2006integratingmmd}, while \cite{sun2016return, sun2016deep, peng2019momentm3sda} introduce metrics based on second- or higher-order statistics. 
Another popular line learns domain-invariant representation using adversarial training.  It has been widely studied \cite{cui2020gradually, chen2020adversarialadla, long2018cdane, tzeng2017adversarialadda, sankaranarayanan2018generatetoadapt, chen2018reweighted, volpi2018adversarialfeature, liu2019transferabletat, saito2018maximummcd, lu2020stochastic} since the seminal work DANN \cite{ganin2015dann, ganin2016domaindann}. In general, a domain discriminator is trained to distinguish the source domain from the target domain, meanwhile a feature extractor is trained to fool the discriminator to arrive at aligned features. SymNets\cite{zhang2019domainsymnet} designs symmetric object classifiers which also play a role of domain discriminator. CDAN \cite{long2018cdane} conditions the adversarial model on the discriminative information conveyed in the classifier predictions. MCD \cite{saito2018maximummcd} and STAR \cite{lu2020stochastic} build an adversarial framework to reduce the domain gap measured by the collision of two reduplicated object classifiers. 
GVB \cite{cui2020gradually} balances adversarial training via constructing bridge layers on both the generator and discriminator. 

These approaches all directly optimize domain alignment and classification tasks, while ignoring the optimization inconsistency between these two objectives. 
We propose a meta-optimization based strategy to mitigate the inconsistency for better UDA.

\noindent
\textbf{UDA for Object Detection.} Learning domain adaptive deep object detector was first studied in DA-Faster \cite{chen2018domaindafaster}, where they perform image-level and instance-level alignments via adversarial leaning. SW-DA \cite{saito2019strongweak} reduces domain gap via aligning global-level and local-level features. 
Zhu \etal \cite{zhu2019adaptingselectivealignment} propose to align domains at regions clustered by K-means. 
EPM \cite{hsu2020everyepm} proposes the center-aware alignment based on the center map generated by an anchor-free detector \cite{tian2019fcos}. Other variants exploit style-transfer \cite{kim2019selfwstbsr}, progressive alignment \cite{inoue2018crosswatercolor},  and hierarchical alignment\cite{zhuang2020ifan}. 

We also validate the effectiveness of our proposed MetaAlign on UDA for object detection task.

\noindent
\textbf{Neural Network Meta Learning.} Meta-learning \cite{schmidhuber1995learning, thrun2012learning}  (a.k.a. learning to learn) has a long standing history.  
Recently, it has been widely applied to the optimization of deep neural networks \cite{andrychowicz2016learning, li2017learning} and few-shot classification \cite{koch2015siamese, vinyals2016matching}. Model-Agnostic Meta-Learning (MAML) \cite{finn2017maml} is proposed for few-shot learning and reinforcement learning, which aims to find good parameters initialization for fast adaptation to new tasks. \cite{li2017learningmldg, qiao2020learningsingledomain, balaji2018metareg} introduce meta-learning to Domain Generalization (DG) to synthesize the source-target domain shift during training.
Li \etal \cite{li2020onlineMetaMSDA} adopt MAML to provide better initialization condition for Multi-Source Domain Adaptation.

In this work, we pinpoint the underlying optimization inconsistency of the domain alignment objective and classification objective used for UDA. We are the first to mitigate it via a meta-optimization based strategy by treating the two objectives as meta-train and meta-test tasks respectively. 

\noindent\textbf{Domain Generalization.}
In contrast to UDA, Domain Generalization (DG) is applied in a more challenging scenario where target domain is inaccessible during training \cite{muandet2013domainfirstdg}. One category of methods for DG attempts to learn domain-invariant features \cite{ghifary2015domainmtae,li2018domainMMDAAE,matsuura2020domainmmld}, which borrows ideas from UDA. 
Li \etal \cite{li2018domainMMDAAE} incorporate MMD as a constraint into the training of an adversarial autoencoder. Ghifary \etal \cite{ghifary2015domainmtae} designs multi-task domain-specific decoders to help the training of the domain-invariant encoder.
Matsuura \etal \cite{matsuura2020domainmmld} use adversarial training to learn features invariant among predicted latent domains.
Other categories exploit data augmentation \cite{shankar2018generalizingcrossgrad, volpi2018generalizingadvaug, qiao2020learningmada}, meta-learning \cite{li2017learningmldg, balaji2018metareg, dou2019domainmasf}, and auxiliary tasks \cite{li2019episodic}.

Our MetaAlign can be applied to the first category of approaches for addressing the optimization inconsistency between domain alignment and classification.

\section{Proposed MetaAlign for UDA} \label{preliminaries}

\noindent
\textbf{Problem Formulation:} Unsupervised Domain Adaptation (UDA) aims to transfer the knowledge from labeled source domain to the unlabeled target domain.
We mainly focus on object classification task but also investigate it for object detection. Without loss of generality, we take classification task as the instantiation to describe
our approach. 

For UDA classification, we denote the source domain as $\mathcal{D}_{\mathcal{S}} = \{(\mathbf{x}_i^s, \mathbf{y}_i^s)\}_{i=1}^{N_{s}}$ with $N_s$ labeled samples, 
where $\mathbf{x}_i^s$ and $\mathbf{y}_i^s$ denote the $i^{th}$ sample and its class label respectively. 
The target domain is denoted as $\mathcal{D}_{\mathcal{T}} = \{\mathbf{x}_i^s\}_{i=1}^{N_{t}}$ with $N_t$ unlabeled samples. Both domains share the same label space $Y=\{1,2,\cdots,K\}$ with $K$ object classes. UDA is expected to train model on $\mathcal{D}_{\mathcal{S}}$ and $\mathcal{D}_{\mathcal{T}}$, to obtain high accuracy on the target test set.

The mainstream UDAs aim to align the source and target domains to alleviate the domain gap. Such alignments are, in general, not designed specially for classification task, \ieno, their optimization may not work harmoniously with that of object classification task. As a result, it may damage the discriminative power of features and thus impede attaining higher performance.
To address this, as illustrated in Fig.~\ref{fig:pipeline}, we introduce a meta-optimization based strategy MetaAlign to encourage the optimization consistency between domain alignment and object classification task itself. 

To be self-included, we first describe several representative domain alignment based UDAs which we use as our baselines.
Then, we introduce our MetaAlign to alleviate the above-mentioned optimization inconsistency problem.

\subsection{Recap of Alignment Based UDAs}
\label{sec: recap}

\begin{figure}[t]
	\centering
	\includegraphics[width=0.47\textwidth]{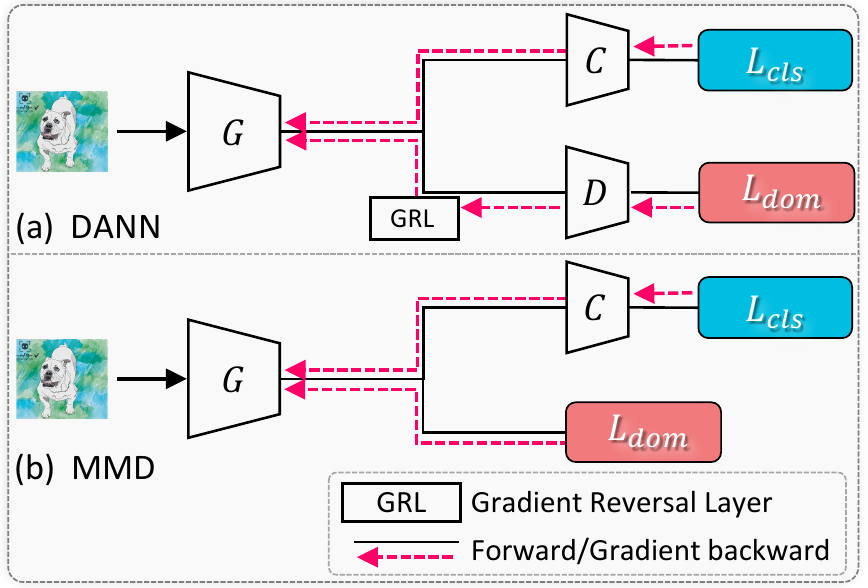}
	\vspace{-0.2cm}
	\caption{Two representative domain alignment based UDA frameworks. (a) DANN. (b) MMD.}
	\label{fig: baselines}
\end{figure}

\begin{figure*}[t]
	\centering
	\includegraphics[width=0.99\textwidth]{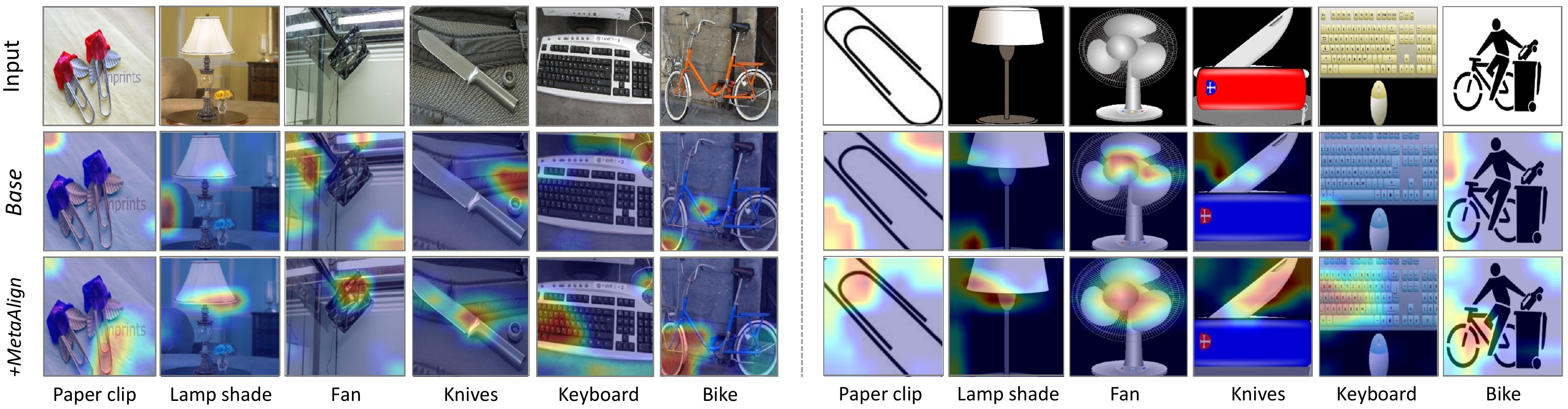}
	\vspace{-0.2cm}
	\caption{Visualization of the Grad-CAMs \cite{selvaraju2017grad} w.r.t. the domain alignment task (domain classifier). 
		The first row of left/right panels show the samples from source (Rw)/target (Cl) domains on Office-Home, while the second and third rows show the Grad-CAMs for Baseline (\textit{Base}) and Baseline with MetaAlign (\textit{+MetaAlign}), respectively.}
	\label{fig: grad_cam}
\end{figure*}

Domain alignment based UDAs include adversarial training based \cite{ganin2015dann, ganin2016domaindann, long2018cdane} and explicit distribution similarity metric based \cite{long2015learning,long2016unsupervised, peng2019momentm3sda, sun2016return}. The core idea of the former category is to train a domain discriminator to distinguish source domain features from target domain features, meanwhile train the feature network to fool the discriminator to implicitly align domains. The latter one explicitly reduces domain discrepancy w.r.t. distribution discrepancy metrics, like Maximum Mean Discrepancy (MMD) \cite{long2016unsupervised, long2015learning}, moment distance \cite{peng2019momentm3sda}, and second-order correlations \cite{sun2016return}.

\noindent\textbf{Adversarial Domain Adaptation.}
To align the domain distributions via adversarial training, such methods try to optimize  
the object classification loss $\mathcal{L}_{cls}$ and the domain alignment loss $\mathcal{L}_{dom}$, simultaneously. Typically, the network is equipped with
a feature extractor/generator $G$, an object classifier $C$, and a domain discriminator/classifier $D$.

The specific object classification loss in the source domain $\mathcal{D}_{\mathcal{S}}$ is formulated as:
\begin{equation}
	\mathcal{L}_{cls} = \frac{1}{N_{s}} \sum_{i=1}^{N_s} \mathcal{L}_{ce} (C(G(\mathbf{x}_i^s)), \mathbf{y}_i^s).
	\label{l_cls}
\end{equation}
where $\mathcal{L}_{ce}$ is a typical cross-entropy loss. 

The domain discriminator $D$ is a two-class classification module, which aims to distinguish the target domain from source domain based on features from $G$. Adversarially, $G$ aims to generate responses with aligned distributions for two domains to fool $D$. Particularly, the domain classification loss can be formulated as:
\begin{equation}
	\begin{split}
		\mathcal{L}_{{dom}_{cls}} = &-\frac{1}{N_s} \sum_{i=1}^{N_s} \log(D(G(\mathbf{x}_i^s))) \\
		&- \frac{1}{N_t} \sum_{j=1}^{N_t} \log(1 - D(G(\mathbf{x}_j^t))). \\
	\end{split}
	\label{l_domain}
\end{equation}

We define the domain alignment loss as $\mathcal{L}_{dom}\!=\!-\mathcal{L}_{{dom}_{cls}}$. The more inseparable (larger $\mathcal{L}_{{dom}_{cls}}$) two domains are, the smaller domain alignment loss $\mathcal{L}_{dom}$ is.
During training, we train $D$ to maximize $\mathcal{L}_{dom}$, meanwhile $\{G, C\}$ to minimize $\mathcal{L}_{cls}$ and $\mathcal{L}_{dom}$:
\begin{equation}
	\begin{split}
		&\max_D \mathcal{L}_{dom}, \\
		&\min_{G, C} \mathcal{L}_{cls} + \mathcal{L}_{dom},\\
	\end{split}
	\label{naive_total_objective}
\end{equation}
where we ignore the hyper-parameter $\lambda$ (\ieno, $\lambda\mathcal{L}_{dom}$) for balancing two losses, for simplicity. Actually, we keep $\lambda$ same as that of the baselines \cite{long2018cdane, cui2020gradually} in our experiments.

Fig. \ref{fig: baselines} (a) shows a seminal work \textbf{\textit{DANN}} \cite{ganin2015dann, ganin2016domaindann}, which constructs $G$ as a CNN feature extractor.
The extracted features from the two domains are fed to two task branches $C$ and $D$ simultaneously. Gradient Reversal Layer (GRL) \cite{ganin2015dann}, which flips the gradients to $G$ from $D$ 
during gradient back propagation, is used to simplify adversarial training. We also take \textbf{\textit{DANNPE}} \cite{cui2020gradually}, an improved variant of \textit{DANN}, as another strong baseline. It differs from \emph{DANN} in two key aspects
: 1) the input of $D$ is the predicted classification probability; 2) $D$ is prioritized on those easy-to-transfer samples by re-weighting with the entropy of object class prediction. Please see Supplementary for more details about \textit{DANNPE}.

\noindent\textbf{Explicit Domain Alignment.} 
Without introducing additional domain discrimination modules, these methods directly reduce the distribution discrepancy between the features of source domain and target domain w.r.t. some discrepancy measurements/metrics.
\textbf{\textit{MMD}} \cite{borgwardt2006integratingmmd} is a representative distribution discrepancy metric. It has been widely employed as the explicit domain alignment constraint for UDA \cite{gretton2007kernelmmd, long2015learning, long2016unsupervised}. Fig. \ref{fig: baselines} (b) illustrates one UDA framework with MMD constraint. Following \cite{li2018domainMMDAAE}, the domain alignment loss becomes:
\begin{equation}
	\mathcal{L}_{dom} = \mathrm{MMD}(\mathbf{F}^s, \mathbf{F}^t)^2.
	\label{l_mmd}
\end{equation}
where $\mathbf{F}^s$ is the feature distribution of source domain. Please refer to Supplementary for more details.

\subsection{Meta-learning to Align $\mathcal{L}_{cls}$ and $\mathcal{L}_{dom}$} \label{sec: meta_learning_to_align}

Adversarial UDAs 
have brought significant performance improvement on multiple benchmarks. They promote the domain alignment, thus reduce the domain gap and enhance the transferability of the models to target domain. The optimization objective of domain alignment is to reduce discrepancy of features between source domain and target domain. However, without explicit coordination with the task of classification, the optimization direction of alignment may be inconsistent with that of the classification task itself. Such inconsistency could hinder the optimization and lead to inferior performance. 

The optimizations of domain alignment and classification can be considered as two tasks. 
In Fig. \ref{fig: grad_cam}, we use \emph{DANNPE} as our baseline to visualize the Grad-CAMs \cite{selvaraju2017grad}, which \textit{produce visual explanations for decisions and reflect the ``important" region of the input for the predictions} w.r.t. the domain alignment task. The second row (\textit{Base}) denotes the Grad-CAMs obtained from the baseline, 
where the regions with higher responses indicate that their features are indistinguishable to $D$ (considering the GRL has flipped the gradients). 
We can see that \emph{Base} attains alignment usually on regions irrelevant to objects (\egno, backgrounds) or only on small partial regions of objects. 
The features of some foreground objects are still not aligned well, which would impede the transferability of the models on classification. 
It is well known that the foreground object regions are most discriminative for object classification \cite{selvaraju2017grad, zhou2016cam}. 
Not aligning the features of foreground objects well would damage the performance of classification.

With the alignment objective and the classification objective separately assigned to the network, \emph{there is a lack of effective interaction between the domain alignment task and classification task. These two tasks may have different gradient descent directions of their optimizations, resulting in optimization inconsistency.}

\begin{table*}[t]
	% \small
	\renewcommand\arraystretch{1.2}
	\begin{center}
		\resizebox{\textwidth}{!}{    
			\begin{tabular}{l@{}|ccccccccccccc}
				\toprule
				Method& Ar$\rightarrow$Cl & Ar$\rightarrow$Pr & Ar$\rightarrow$Rw & Cl$\rightarrow$Ar & Cl$\rightarrow$Pr & Cl$\rightarrow$Rw & Pr$\rightarrow$Ar & Pr$\rightarrow$Cl & Pr$\rightarrow$Rw & Rw$\rightarrow$Ar & Rw$\rightarrow$Cl & Rw$\rightarrow$Pr & Avg \\
				\hline
				Source-Only \cite{he2016deep} & 34.9 & 50.0 & 58.0 & 37.4 & 41.9 & 46.2 & 38.5 & 31.2 & 60.4 & 53.9 & 41.2 & 59.9 & 46.1 \\
				% 		\rowcolor{Gray} DAN(ICML'15)\cite{long2015learning} & 43.6 & 57.0 & 67.9 & 45.8 & 56.5 & 60.4 & 44.0 & 43.6 & 67.7 & 63.1 & 51.5 & 74.3 & 56.3 \\
				\rowcolor{Gray}MCD(CVPR'18)\cite{saito2018maximummcd}  &48.9	&68.3	&74.6	&61.3	&67.6	&68.8	&57.0	&47.1	&75.1	&69.1	&52.2	&79.6	&64.1 \\
				TAT(ICML'19)\cite{liu2019transferabletat} & 51.6 & 69.5 & 75.4 & 59.4 & 69.5 & 68.6 & 59.5 & 50.5 & 76.8 & 70.9 & 56.6 & 81.6 & 65.8 \\
				\rowcolor{Gray}ALDA(AAAI'20)\cite{chen2020adversarialadla} & 53.7 & 70.1 & 76.4 & 60.2 & 72.6 & 71.5 & 56.8 & 51.9 & 77.1 & 70.2 & 56.3 & 82.1 & 66.6 \\ Sym(CVPR'19)\cite{zhang2019domainsymnet} & 47.7 & 72.9 & 78.5 & 64.2 & 71.3 & 74.2 & 63.6  & 47.6 & 79.4 & 73.8 & 50.8 & {82.6} & 67.2 \\
				\rowcolor{Gray}TADA(AAAI'19)\cite{wang2019transferabletada} & 53.1 & 72.3 & 77.2 & 59.1 & 71.2 & 72.1 & 59.7 & 53.1 & 78.4 & 72.4 & 60.0 & 82.9 & 67.6 \\
				MDD(ICML'19)\cite{zhang2019bridgemdd} &
				54.9 & 73.7 & 77.8 & 60.0 & 71.4 &71.8 & 61.2 & 53.6 & 78.1 & 72.5 & 60.2 & 82.3 & 68.1 \\
				\rowcolor{Gray} BNM(CVPR'20)\cite{cui2020towardsbnm}
				& 56.2 & 73.7 & 79.0 & 63.1 & 73.6 & 74.0 & 62.4 & 54.8 & 80.7 & 72.4 & 58.9 & 83.5 & 69.4 \\
				\hline
				\hline
				MMD & 49.1 & 67.0 & 74.7 & 54.5 & 62.9 & 65.7 & 55.3 & 45.7 & 74.5 & 68.1 & 52.5 & 78.6 & 62.3 \\
				\rowcolor{Gray} +MetaAlign & $49.4_\uparrow$ & $67.2_\uparrow$ & $75.5_\uparrow$ & $58.6_\uparrow$ & $64.7_\uparrow$ & $67.2_\uparrow$ & $55.5_\uparrow$ & $46.1_\uparrow$ & $74.8_\uparrow$ & $69.0_\uparrow$ & $52.1_\downarrow$ & $78.9_\uparrow$ & $63.3_\uparrow$ \\
				\hline
				DANN(ICML'15)\cite{ganin2016domaindann}$\dagger$ & 45.8 & 63.4 & 71.9 & 53.6 & 61.9 & 62.6 & 49.1 & 39.7 & 73.0 & 64.6 & 47.8 & 77.8 & 59.2 \\
				\rowcolor{Gray} +MetaAlign & $48.6_\uparrow$ & $69.5_\uparrow$ & $76.0_\uparrow$ & $58.1_\uparrow$ & $65.7_\uparrow$ & $68.3_\uparrow$ & $54.9_\uparrow$ & $44.4_\uparrow$ & $75.3_\uparrow$ & $68.5_\uparrow$ & $50.8_\uparrow$ & $80.1_\uparrow$ & $63.3_\uparrow$ \\
				\hline
				CDAN(NeurIPS'18)\cite{long2018cdane} & 50.7 & 70.6 & 76.0 & 57.6 & 70.0 & 70.0 & 57.4 & 50.9 & 77.3 & 70.9 & 56.7 & 81.6 & 65.8 \\
				\rowcolor{Gray}+MetaAlign & $55.2_\uparrow$ & $70.5_\downarrow$ & $77.6_\uparrow$ & $61.5_\uparrow$ & $70.0_{-}$ & $70.0_{-}$ & $58.7_\uparrow$ & $55.7_\uparrow$ & $78.5_\uparrow$ & $73.3_\uparrow$ & $61.0_\uparrow$ & $81.7_\uparrow$ & $67.8_\uparrow$ \\
				\hline
				DANNPE  & 54.7 & 72.8 & 78.5 & 62.3 & 71.1 & 73.1 & 61.0 & 53.0 & 80.0 & 72.8 &56.5 & 83.4 &68.3 \\
				\rowcolor{Gray} +MetaAlign & $57.1_\uparrow$ & $74.5_\uparrow$ & $80.1_\uparrow$ & $64.9_\uparrow$ & $73.6_\uparrow$ & $74.6_\uparrow$ & $62.5_\uparrow$ & $54.8_\uparrow$ & $80.6_\uparrow$ & $73.6_\uparrow$ & $60.3_\uparrow$ & $84.7_\uparrow$ & $70.1_\uparrow$ \\
				\hline
				GVB(CVPR'20)\cite{cui2020gradually} & 57.0 & 74.7 & 79.8 & 64.6 & 74.1 & 74.6 & 65.2 &  55.1 & 81.0 & \textbf{74.6} & 59.7 & 84.3 & 70.4 \\
				\rowcolor{Gray} +MetaAlign & $\textbf{59.3}_\uparrow$ & $\textbf{76.0}_\uparrow$ & $\textbf{80.2}_\uparrow$ & $\textbf{65.7}_\uparrow$ & $\textbf{74.7}_\uparrow$ & $\textbf{75.1}_\uparrow$ & $\textbf{65.7}_\uparrow$ & $\textbf{56.5}_\uparrow$ & $\textbf{81.6}_\uparrow$ & $74.1_\downarrow$ & $\textbf{ 61.1 }_\uparrow$ & $\textbf{85.2}_\uparrow$ & $\textbf{71.3}_\uparrow$ \\
				\thickhline
		\end{tabular}}
		\vspace{-0.3cm}
		\caption{
			Classification accuracy (\%) of different UDAs on Office-Home with ResNet-50 as backbone.  We re-implement all the adopted baselines for MetaAlign. $\dagger$ denotes our re-implemented result is different from the one reported in other papers.}
		\label{table:uda_office-home}
	\end{center}
\end{table*}

One natural question to ask arises:
how to easily incorporate the optimization consistency constraint for domain alignment and classification? 

We propose to promote the optimization consistency between these two tasks by exploring a meta-optimization strategy. We draw inspiration from 
Model-Agnostic Meta-Learning (MAML) \cite{finn2017maml}, which generally separates the samples into meta-train splits and meta-test splits, and uses meta-learning to train the model to be learned quickly on meta-test samples given the knowledge in meta-train. 
Meta-Learning Domain Generalization (MLDG) \cite{li2017learningmldg} simulates training-test domain shift during training by synthesizing virtual test domain, with meta-optimization objective requiring that steps to improve training domain performance should also improve testing domain performance. 

In our work, we leverage meta-optimization to coordinate the domain alignment task and classification task. Particularly, rather than splitting the samples into meta-train and meta-test as in \cite{finn2017maml,li2017learningmldg}, \textbf{we treat the domain alignment task and classification task as meta-train (or meta-test) and meta-test (or meta-train) for the same set of samples}. 

\begin{algorithm}[t]
	\small
	\begin{algorithmic}[1]
		\State \textbf{Input}: Source and target data sets  $\mathcal{D}_\mathcal{S}$ and  $\mathcal{D}_\mathcal{T}$
		\State \textbf{Init}: parameters $\Psi = \{ \theta, \phi_c\, \beta, \phi_d\}$, learning rate $\eta, \alpha$
		\For{t \textbf{in} iterations}
		\Statex \textbf{\phantom{for }Meta-train:}
		\State Compute domain alignment loss $\mathcal{L}_{dom}$ \Comment{Eq. (\ref{l_domain})}
		\State Update $\theta$ w.r.t. $\mathcal{L}_{dom}$: 
		\Statex \begin{center}
			$\theta^{t+1}_m \leftarrow \theta^t_m - \alpha\beta_m\nabla_{\theta_m^t}\mathcal{L}_{dom}(\theta^t, \phi_d^t)$
		\end{center}
		\Statex \textbf{\phantom{for }Meta-test:}
		\State Compute classification loss $\mathcal{L}_{cls}(\theta^{t+1}, \phi_c^t)$ \Comment{Eq. (\ref{l_cls})}
		\Statex \textbf{\phantom{for }Meta optimization:}
		\State Compute total loss $\mathcal{L}_{total}$ \Comment{Eq. (\ref{original_meta_weights})}
		\State Update model parameters: 
		\Statex \begin{center}
			$\Psi^{t+1}\leftarrow \Psi^{t} - \eta\bigtriangledown_{\Psi^t}\mathcal{L}_{total}$ 
		\end{center}
		\EndFor
		\State \textbf{Output}: $\theta, \phi_c$
	\end{algorithmic}
	\caption{MetaAlign Optimization Algorithm}
	\label{alg:algorithm}
\end{algorithm}

A UDA network is jointly optimized with classification objective and domain alignment objective. The learnable network parameters consists of shared parameters $\theta$, the parameters specific to domain alignment $\phi_d$, and the parameters specific to classification $\phi_c$. The general optimization objective can be formulated as:
\begin{equation}
	\operatorname*{min}_{\theta,\phi_c} \operatorname*{max}_{\phi_d} \mathcal{L}_{dom}(\theta,\phi_d) + \mathcal{L}_{cls}(\theta,\phi_c).
	\label{objective}
\end{equation}
which does not handle the potential optimization inconsistency of the two tasks.

With the intuition that meta-test task (\egno, classification) will be used to evaluate the effect of the model optimization on meta-train task (\egno, domain alignment), the overall meta-optimization objective can be formulated as:
\begin{equation}
	\operatorname*{min}_{\theta,\phi_c}\operatorname*{max}_{\phi_d} \mathcal{L}_{dom}(\theta,\phi_d) + \mathcal{L}_{cls}(\theta - \alpha \nabla_\theta \mathcal{L}_{dom}(\theta, \phi_d),\phi_c).
	\label{original_meta}
\end{equation}
which aims to optimize both the loss of meta-train $\mathcal{L}_{dom}$, and that of meta-test $\mathcal{L}_{cls}$ 
after updating $\theta$ during meta-train by one gradient descent step: $\theta' \leftarrow \theta-\alpha \nabla_\theta \mathcal{L}_{dom}(\theta, \phi_d)$, where $\alpha$ denotes the \textit{meta-learning rate}. To alleviate the computational complexity, similar to \cite{dou2019domainmasf, finn2017maml}, we omit higher-order ones during the back-propagation of gradients.

This meta-optimization enables the explicit interaction between the two tasks. Following  \cite{li2017learningmldg}, we analyse Eq. (\ref{original_meta}) by approximating the second term using its first-order Taylor expansion as:
\begin{equation}
	\begin{split}
		\operatorname*{min}_{\theta,\phi_c}\operatorname*{max}_{\phi_d}& ~\mathcal{L}_{dom}(\theta,\phi_d) + \mathcal{L}_{cls}(\theta,\phi_c) \\
		& - \alpha\nabla_\theta \mathcal{L}_{cls}(\theta,\phi_c) \nabla_\theta\mathcal{L}_{dom}(\theta, \phi_d).
	\end{split}
	\label{taylor_meta}
\end{equation}

Compared with the general optimization objective Eq. (\ref{objective}), the additional last term in Eq. (\ref{taylor_meta}) 
maximizes the dot product of 
$\nabla_\theta \mathcal{L}_{cls}$ 
and $\nabla_\theta\mathcal{L}_{dom}$, which encourages the consistency of the optimization directions (gradients) of two tasks. \emph{In this way, both domain alignment and object classification are optimized in a coordinated way.} We refer to our method as \textit{MetaAlign} which \textit{Align}s the domain alignment task and classification task with a \textit{Meta}-optimization strategy.

In Eq. (\ref{taylor_meta}), we place the consistency constraint on all the parameters $\theta$ (including different layers) shared by two tasks. 
Actually, different layers in a CNN learn features with different semantics. Intuitively, they should be treated differently when aligning their optimizations, where the optimization consistency for some layers may be more important than other layers. We propose to adaptively learn the importance levels for different groups of layers for better optimization. Particularly, we partition the layers into $M$ groups (\egno, each convolutional block of ResNet as a group) and learn a scalar weight $\beta_m$ for the $m^{th}$ group. The optimization objective is thus formulated as:
\begin{equation}
	\begin{split}
		&\operatorname*{min}_{\theta, \phi_c, {\beta}}\operatorname*{max}_{\phi_d} ~ \mathcal{L}_{dom}(\theta,\phi_d) \\ 
		&\phantom{arg}+ \mathcal{L}_{cls}\left(\left\{\theta_{m} - \alpha \beta_{m} \nabla_{\theta_m} \mathcal{L}_{dom}(\theta, \phi_d)\right\}_{m=1}^{M}, \phi_c\right)\\
		&\phantom{arg}+ \mathcal{L}_\beta(\beta).\\
	\end{split}
	\label{original_meta_weights}
\end{equation}
where $\theta_m$ denotes the parameters of the $m^{th}\in\{1, \dots, M\}$  group. To avoid trivial solution, we add $L_1$ constraint on $\beta=\{\beta_m\}_m$: $\mathcal{L}_{\beta} = ||\sum_{m=1}^{M}\beta_m - B||_1$, where $B$ is a hyper-parameter. We denote the entire losses in Eq. (\ref{original_meta_weights}) as $\mathcal{L}_{total}$ for simplicity.

\begin{table*}[t]
	\renewcommand\arraystretch{1.2}
	\centering
	% 	\small
	\resizebox{0.78\textwidth}{!}{
		\begin{tabular}{l|ccccccc}
			\hline
			Method& A $\rightarrow$ W & A $\rightarrow$ D & W $\rightarrow$ A & W $\rightarrow$ D & D $\rightarrow$ A & D $\rightarrow$ W  & Avg \\
			\hline
			Source-Only \cite{he2016deep} & 68.4$\pm$0.2 & 68.9$\pm$0.2 & 60.7$\pm$0.3 & 99.3$\pm$0.1 & 62.5$\pm$0.3 & 96.7$\pm$0.1 & 76.1 \\
			% 		\rowcolor{Gray}DAN (ICML'15) \cite{long2015learning} & 80.5$\pm$0.4 & 78.6$\pm$0.2 & 62.8$\pm$0.2 & 99.6$\pm$0.1 & 63.6$\pm$0.3 & 97.1$\pm$0.2 & 80.4 \\
% 			\rowcolor{Gray}DANN (ICML'15) \cite{ganin2015dann} & 82.0$\pm$0.4 & 79.7$\pm$0.4 & 67.4$\pm$0.5 & 99.1$\pm$0.1 & 68.2$\pm$0.4 & 96.9$\pm$0.2 & 82.2 \\
% 			MCD (CVPR'18) \cite{saito2018maximummcd} & 88.6$\pm$0.2 & 92.2$\pm$0.2 & 69.7$\pm$0.3 & \textbf{100.0}$\pm$.0 & 69.5$\pm$0.1 & 98.5$\pm$0.1 & 86.5 \\
			\rowcolor{Gray}CDAN (NeurIPS'18) \cite{long2018cdane} & 94.1$\pm$0.1 & 92.9$\pm$0.2 & 69.3$\pm$0.3 & \textbf{100.0}$\pm$.0 & 71.0$\pm$0.3 & 98.6$\pm$0.1 & 87.7 \\
			TAT (ICML'19) \cite{liu2019transferabletat} &
			92.5$\pm$0.3 & 93.2$\pm$0.2 & 72.1$\pm$0.3 & \textbf{100.0}$\pm$.0 & 73.1$\pm$0.3 & \textbf{99.3}$\pm$0.1 & 88.4\\
			\rowcolor{Gray}TADA (AAAI'19) \cite{wang2019transferabletada} &
			94.3$\pm$0.3 & 91.6$\pm$0.3 & 73.0$\pm$0.3 & 99.8$\pm$0.2 & 72.9$\pm$0.2 & 98.7$\pm$0.1 & 88.4 \\
			Sym (CVPR'19) \cite{zhang2019domainsymnet} &
			90.8$\pm$0.1 & 93.9$\pm$0.5 & 72.5$\pm$0.5 & \textbf{100.0}$\pm$.0 & 74.6$\pm$0.6 & 98.8$\pm$0.3 & 88.4 \\
			% 		Sym (CVPR'19) \cite{zhang2019domainsymnet} & 90.8$\pm$0.1 & 93.9$\pm$0.5 & 72.5$\pm$0.5 & 100.0$\pm$.0 & 74.6$\pm$0.6 & 98.8$\pm$0.3 & 88.4 \\
			\rowcolor{Gray}BNM (CVPR'20) \cite{cui2020towardsbnm} & 92.8 & 92.9 & 73.8 & \textbf{100.0} & 73.5 & 98.8 & 88.6 \\
			ALDA (AAAI'20) \cite{chen2020adversarialadla} & \textbf{95.6}$\pm$0.5 & 94.0$\pm$0.4 & 72.5$\pm$0.2 & \textbf{100.0}$\pm$.0 & 72.2$\pm$0.4 & 97.7$\pm$0.1 & 88.7 \\
			\rowcolor{Gray}MDD (ICML'19) \cite{zhang2019bridgemdd} & 94.5$\pm$0.3 & 93.5$\pm$0.2 & 72.2$\pm$0.1 & \textbf{100.0}$\pm$.0 & 74.6$\pm$0.3 & 98.4$\pm$0.1 & 88.9 \\
			\hline
			\hline
			DANNPE & 92.5$\pm$0.5 & 89.9$\pm$0.3 & 71.1$\pm$0.3 & 99.9$\pm$0.1 & 70.7$\pm$0.4 & 98.5$\pm$0.3 & 87.0 \\
			\rowcolor{Gray}+MetaAlign & $93.9\!\pm\!0.4_\uparrow$ & $91.6\!\pm\!0.3_\uparrow$ & $\textbf{74.1}\!\pm\!0.2_\uparrow$ & $\textbf{100.0}\!\pm\!.0_\uparrow$ & $73.7\!\pm\!0.2_\uparrow$ & $98.7\!\pm\!0.2_\uparrow$ & $88.7_\uparrow$ \\
			\hline
			GVB (CVPR'20) \cite{cui2020gradually}$\dagger$ & 92.0$\pm$0.3 & 91.4$\pm$0.5 & 73.4$\pm$0.1 & \textbf{100.0}$\pm$.0 & 74.9$\pm$0.5 & 98.7$\pm$.0 & 88.3\\
			\rowcolor{Gray}+MetaAlign & $93.0\!\pm\!0.5_\uparrow$ & $\textbf{94.5}\!\pm\!0.3_\uparrow$ & $73.6\!\pm\!.0_\uparrow$ & $\textbf{100.0}\!\pm\!.0_{-}$ & $\textbf{75.0}\!\pm\!0.3_\uparrow$ & $98.6\!\pm\!.0_\downarrow$ & $\textbf{89.2}_\uparrow$\\
			\hline
	\end{tabular}}
	\vspace{-0.3cm}
	\caption{
		Classification accuracy (mean $\pm$ std \%) of different UDAs on Office31 with ResNet-50 as backbone. We re-implement all the adopted baselines. $\dagger$ denotes the result is different from the reported one in the original paper.}
	\label{table:uda_office31}
\end{table*}

\noindent
\textbf{Training.} In practice, we can iteratively choose one of the two tasks (domain alignment task and object classification task) as meta-train while the other as meta-test. We describe the training procedure in Alg.~\ref{alg:algorithm}. For simplicity, we only show one case, \ieno, the domain alignment task and object classification task are taken as meta-train and meta-test respectively. The other case is similar.

\section{Experiments}

In this section, we validate the effectiveness of our proposed MetaAlign on UDA for classification in Sec. \ref{subsec:classification} and for object detection in Sec.~\ref{subsec:detection}. Besides, we further show its effectiveness on DG for classification in Sec.~\ref{subsec:DG}. Due to the space constraint, we refer readers to Supplementary for more details for all experiments.

\subsection{UDA for Classification}
\label{subsec:classification}

\subsubsection{Datasets and Settings} 
We conduct UDA experiments on two popular benchmarks \textbf{Office31} \cite{saenko2010adaptingoffice31} and \textbf{Office-Home} \cite{venkateswara2017deep-officehome}. 1) \textbf{Office31} is a standard benchmark for domain adaptative classification. It contains images of 31 categories, drawn from three domains: Amazon (A), Webcam (W), and DSLR (D). Following the typical setting \cite{cui2020gradually, zhang2019domainsymnet, saito2018maximummcd}, we evaluate the methods on one-source to one-target domain adaptation.
\begin{table}[t]
	\centering
	\renewcommand{\arraystretch}{1.2}
	\begin{adjustbox}{width=0.42\textwidth}
		\begin{tabular}{@{}c|ccc}
			\hline
			Source-Only & DANNPE & +MetaAlign w/o $\beta$ & +MetaAlign \\
			\hline
			46.1 & 68.3 & 69.7 & 70.1 \\
			\hline
		\end{tabular}
	\end{adjustbox}
	\vspace{-0.3cm}
	\caption{Ablation study on Office-Home with DANNPE as the baseline.}
	\label{tab:ablation_study}
\end{table}
2) \textbf{Office-Home} is a more challenging recent dataset for UDA. It consists of images from 4 different domains: Art (Ar), Clip Art (Cl), Product (Pr), and Real-World (Rw). Each domain contains 65 object categories found typically in office and home environments. We evaluate our method in all the 12 one-source to one-target adaptation cases. All reported results are obtained from the average of multiple runs.

\begin{figure}[t]
	\centering
	\includegraphics[width=0.48\textwidth]{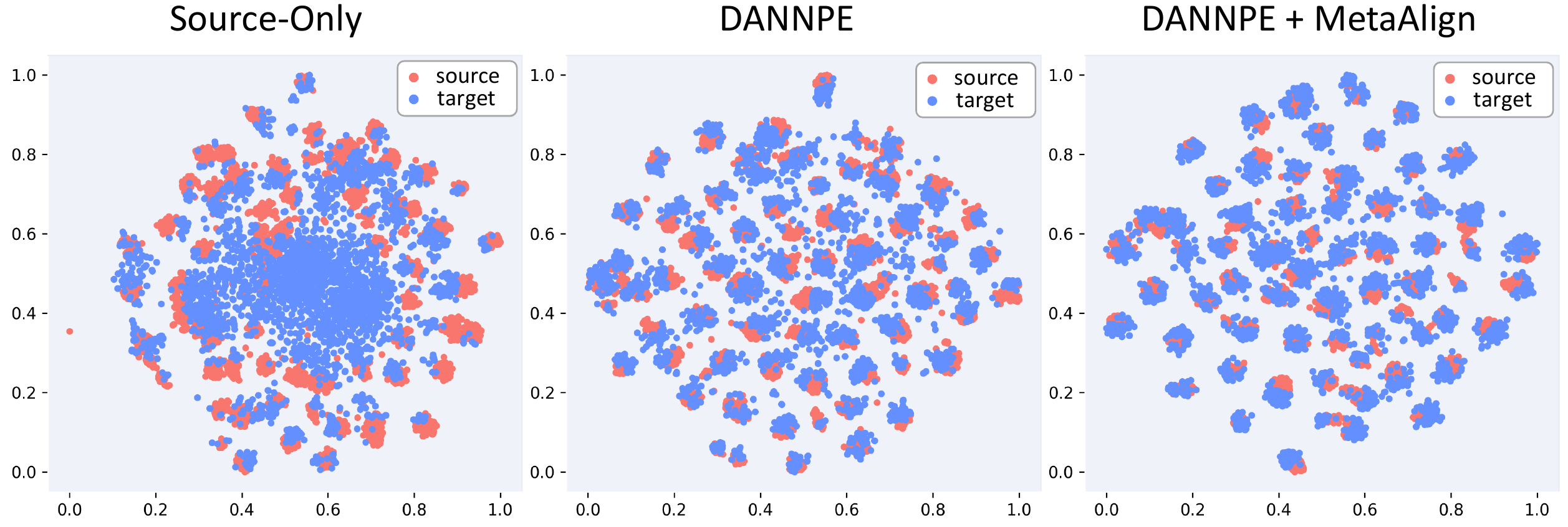}
	\vspace{-0.5cm}
	\caption{t-SNE visualization of features learned by Source-Only (left), DANNPE (middle) and DANNPE+MetaAlign (right).
	}
	\label{fig: viz_tsne_oh}
\end{figure}

\begin{table*}[t]
	\renewcommand\arraystretch{1.2}
	\begin{center}
		\resizebox{0.97\textwidth}{!}{    
			\begin{tabular}{c|ccccccccc|cc} % 11
				\toprule
				Source $\rightarrow$ Target & AGG & \makecell{MMD-AAE\cite{li2018domainMMDAAE}\\(CVPR'18)}
				% 		& MAML\cite{finn2017maml} 
				&  \makecell{CrossGrad\cite{shankar2018generalizingcrossgrad}\\(ICLR'18)} & \makecell{MetaReg\cite{balaji2018metareg}\\(NeurIPS'18)} &  \makecell{JiGen\cite{carlucci2019domainjigen}\\(CVPR'19)} &	\makecell{MLDG\cite{li2017learningmldg}\\(AAAI'18)} &					\makecell{MASF\cite{dou2019domainmasf}\\(NeurIPS'19)} & \makecell{Epi-FCR\cite{li2019episodic}\\(ICCV'19)} & \makecell{MMLD\cite{matsuura2020domainmmld}\\(AAAI'20)} & DANNPE & +MetaAlign \\
				\hline
				\rowcolor{white}
				A,C,S~ $\rightarrow$~~ P~~ ~~ & 94.4 & 96.0 & 94.0 & 94.3 & 96.0 & 94.3 &  95.0 & 93.9 & \textbf{96.1} & 95.2 & 95.5 \\
				\rowcolor{Gray}
				C,P,S~~ $\rightarrow$ ~~A~~ ~~& 77.6 & 75.2 & 78.7 & 79.5 & 79.4 & 79.5 & 80.3 & \textbf{82.1} & 81.3 & 76.5 & 78.5 \\
				A,P,S~~ $\rightarrow$~~ C~~ ~~ & 73.9 & 72.7 &  73.3 & 75.4 & 77.3 & 75.3 & 77.2 & 77.0 & 77.2 & 77.2 & \textbf{77.8} \\
				\rowcolor{Gray}
				A,C,P ~$\rightarrow$~~ S~~ ~~ & 70.3 & 64.2 & 65.1 & 72.2 & 71.4 & 71.5 & 71.7 & 73.0 & 72.3 & 74.1 & \textbf{75.7} \\
				\hline
				Avg. & 79.1 & 77.0 & 77.8 & 80.4 & 80.5 & 80.7 & 81.0 & 81.5 & 81.8 & 80.7 &\textbf{ 81.9} \\
				\hline
		\end{tabular}}
		\vspace{-0.2cm}
		\caption{
			Accuracy (\%) of different domain generalization methods on PACS with ResNet-18 as backbone. Best in bold.}
		\label{table:dg_pacs}
	\end{center}
\end{table*}

\begin{table}[t]
	\centering
	\renewcommand{\arraystretch}{1.2}
	\begin{adjustbox}{width=0.48\textwidth}
		\begin{tabular}{c|ccccccc}
			\toprule
			Methods & bike & bird & car & cat & dog & person & mAP \\ 
			\hline
			Source Only & 68.8 & 46.8 & 37.2 & 32.7 & 21.3 & 60.7 & 44.6 \\
			\rowcolor{Gray}\Gape[0pt][0pt]{\makecell{BDC-Faster\cite{saito2019strongweak}(CVPR'19)}}
			& 68.6 & 48.3 & 47.2 & 26.5 & 21.7 & 60.5 & 45.5 \\
			\makecell{WST+BSR\cite{kim2019selfwstbsr}(ICCV'19)}
			& 75.6 & 45.8 & \textbf{49.3 }& 34.1 & 30.3 & 64.1 & 49.9 \\
			\rowcolor{Gray}\Gape[0pt][0pt]{\makecell{MAF\cite{he2019multimaf}(ICCV'19)}}
			& 73.4 & 55.7 & 46.4 & 36.8 & 28.9 & 60.8 & 50.3 \\
			\makecell{DT-UDA\cite{inoue2018crosswatercolor}(CVPR'18)}
			& 82.8 & 47.0 & 40.2 & 34.6 & 35.3 & 62.5 & 50.4 \\
			\rowcolor{Gray}\Gape[0pt][0pt]{\makecell{ATF\cite{he2020domainatf}(ECCV`20)}}
			& 78.8 & \textbf{59.9} & 47.9 & \textbf{41.0} & 34.8 & 66.9 & 54.9 \\
			\hline
			\makecell{W-DA\cite{saito2019strongweak}(CVPR'19)} 
			& 66.4 & 53.7 & 43.8 & 37.9 & 31.9 & 65.3 & 49.8 \\
			\rowcolor{Gray}\Gape[0pt][0pt]{W-DA+MetaAlign}
			& $74.9_\uparrow$ & $54.0_\uparrow$ & $43.7_\downarrow$ & $38.1_\uparrow$ & $35.2_\uparrow$ & $66.9_\uparrow$ & $52.1_\uparrow$ \\
			\hline
			\makecell{SW-DA\cite{saito2019strongweak}(CVPR'19)}
			& 76.1 & 52.7 & 49.1 & 36.3 & 40.2 & 66.3 & 53.5 \\
			\rowcolor{Gray}\Gape[0pt][0pt]{SW-DA+MetaAlign}
			& $\textbf{83.7}_\uparrow$ & $53.2_\uparrow$ & $48.7_\downarrow$ & $38.7_\uparrow$ & $\textbf{42.0}_\uparrow$ & $\textbf{67.2}_\uparrow$ & $\textbf{55.6}_\uparrow$ \\
			\thickhline
		\end{tabular}
	\end{adjustbox}
	\vspace{-0.2cm}
	\caption{Performance of UDAs for object detection from Pascal VOC to Watercolor2k in terms of mAP. Best in bold.}
	\label{tab:uda_voc_to_watercolor}
\end{table}

\subsubsection{Ablation Study} 

\noindent\textbf{Effectiveness of MetaAlign on Various Baselines.} 
Our proposed MetaAlign is generic which could be applied to alleviate the optimization inconsistency of most existing domain alignment based UDAs. We use various alignment-based UDAs as our baselines to validate the effectiveness of MeanAlign. Specifically, we adopt five baselines. 1) \textbf{\textit{MMD}}, 2) \textbf{\textit{DANN}}, and 3) \textbf{\textit{DANNPE}} have been described in Sec. \ref{sec: recap} in detail. 4) \textbf{\textit{CDAN}} \cite{long2018cdane} aligns domain at class-level. 5) \textbf{\textit{GVB}} \cite{cui2020gradually} is a most recent state-of-the-art method with enhanced $C$ and $D$. Note that \emph{MMD} explicitly reduces the discrepancy of domains, while the others belong to adversarial learning based approaches.

Table \ref{table:uda_office-home} shows the comparisons on Office-Home. Our MetaAlign consistently improves the accuracy of all the five baselines, \ieno, 1.0\%, 3.9\%, 2.0\%, 1.8\%, 0.9\% on average for \emph{MMD}, \emph{DANN}, \emph{CDAN}, \emph{DANNPE}, \emph{GVB}, respectively, regardless of the design differences on $\mathcal{L}_{dom}$. With the help of MetaAlign, domain alignment and classification are optimized in a coordinated way, resulting in more efficient optimization.

\noindent\textbf{Effectiveness of Re-weighting with $\beta$ in MetaAlign.} 
In our design, $\beta_m, m=1,\cdots, M$ in Eq.~(\ref{original_meta_weights}) are learned to allocate the levels of importance for different layer groups on the consistency constraint. We validate its effectiveness on top of the baseline \emph{DANNPE} on Office-Home in Table~\ref{tab:ablation_study}.
Our MetaAlign  improves over \emph{DANNPE} by 1.8\% in accuracy. Without $\beta$, the gain decreases to 1.4\%. 

\noindent\textbf{Which Task as Meta-Train?} 
We could treat any one of the two tasks as meta-train and the other as meta-test.  We could also iteratively exchange their roles during training. Experimental results show their results are very close ($<$ 0.3\% accuracy). The explanation lies in that these settings have the same optimization objective as Eq. (\ref{taylor_meta}).

\subsubsection{Comparisons with State-of-the-Arts} 

To compare with previous state-of-the-art UDAs, we incorporate our MetaAlign optimization strategy into the recent strong UDA method \emph{GVB} \cite{cui2020gradually}, termed as \emph{GVB+MetaAlign}. Table~\ref{table:uda_office-home} and Table~\ref{table:uda_office31} show the comparisons with the state-of-the-art approaches on Office-Home and Office31, respectively. \emph{GVB+MetaAlign} outperforms \emph{GVB} and achieves the best performance on both datasets. 

\subsubsection{Feature Visualization}

As analysed in Sec. \ref{sec: meta_learning_to_align}, we expect MetaAlign to enforce the domain alignment task and object classification task to be optimized in a coordinated way. To validate it, we visualize the Grad-CAMs \cite{selvaraju2017grad} w.r.t. the domain alignment task in Fig.~\ref{fig: grad_cam}. With MetaAlign, the domain alignment task focuses on the regions more related to foreground objects compared with Baseline. These regions play the most important role for object classification task \cite{zhou2016cam, selvaraju2017grad}, which is also validated in the Supplementary. Aligning these features helps improve object classification accuracy indeed.

We also visualize the learned features by t-SNE \cite{saito2019strongweak}, on task Pr $\rightarrow$ Cl in Fig. \ref{fig: viz_tsne_oh}. It is shown that Source-Only works well only in source domain but poorly in target domain without domain alignment. \textit{DANNPE} aligns domains well via adversarial learning. Further, employing MetaAlign arrives at much better alignment results, where the clusters are more compact and less data points scatter at the boundaries between clusters. The visualization result further validates the effectiveness of MetaAlign for domain alignment based UDA.

\subsection{Experiments on UDA for Object Detection}
\label{subsec:detection}

Adversarial learning has also been exploited in UDA for object detection \cite{chen2018domaindafaster, saito2019strongweak}. It is natural to align foreground objects across domains in this task, where the aforementioned optimization inconsistency issue still exists. Our MetaAlign is generic and is expected to work well for this task. We take a Faster RCNN\cite{ren2015fasterrcnn} based SW-DA \cite{saito2019strongweak} as the baseline, and conduct UDA experiments from Pascal VOC \cite{pascalvoc2007, pascalvoc2012} to Watercolor2k \cite{inoue2018crosswatercolor}. All reported results are obtained from the average of multiple runs. Please refer to Supplementary for details about datasets, experimental settings, and the introduction of competitors.
As shown in Table \ref{tab:uda_voc_to_watercolor}, our MetaAlign strategy improves the mAP of the two baselines \textit{W-DA} (\textit{SW-DA} without local alignment) and \textit{SW-DA} by 2.3 (4.6\%) and 1.9 (3.5\%), respectively. The latter one achieves state-of-the-art performance, compared with recent methods. 
Note that the results on `bird' are unstable due to the insufficient data (the number of bird bounding boxes are about 2.6\% of all bounding boxes in the dataset).
Some qualitative comparisons (see Supplementary) demonstrate MetaAlign improves the object classification accuracy of the predicted bounding boxes, thanks to the coordination between domain alignment and object classification. 
These validate that our MetaAlign strategy is compatible with UDAs for different vision tasks.

\subsection{Experiments on Domain Generalization}
\label{subsec:DG}

Learning domain-invariant features is also widely explored in DG. Therefore we further apply our MetaAlign on DG to validate its generalizability. \textbf{\textit{DANNPE}} is originally designed for UDA. However, its goal, to learn domain-invariant features, fits well with DG. Therefore we repurpose it as a baseline for DG to incorporate our MetaAlign. We perform experiments on PACS \cite{li2017deeperpacs}. All reported results are obtained from the average of multiple runs. Please refer to Supplementary for more details about the dataset, settings, and the description of competitors.	
As shown in Table \ref{table:dg_pacs}, \textit{DANNPE} also works for DG with about 1.6\% improvement over \textit{AGG}. With our proposed MetaAlign strategy, the accuracy is further improved by 1.2\%. We reckon that MetaAlign enforces the domain-invariant features and classification discriminative features to be learned in concert. \textit{DANNPE+MetaAlign} outperforms previous state-of-the-art methods on the average accuracy, especially in the most challenging scenario where the target domain is \textit{Sketch}.
\textit{Sketch} has the largest domain gap with other domains, therefore, more powerful domain-invariant features are required.

\section{Conclusion}

In this paper, we pinpoint the optimization inconsistency problem between the domain alignment task and the classification task itself in alignment-based UDAs. To mitigate it, we propose a meta-optimization based strategy named MetaAlign, which treats one of these two tasks as meta-train and the other as meta-test. The analysis of the optimization objective of MetaAlign reveals that the two tasks will be optimized in a coordinated way.
The experimental results validate that MetaAlign is applicable to various alignment-based UDAs for classification and detection. 

\section{Acknowledgments}
This work was supported in part by NSFC under Grant U1908209, 61632001 and the National Key Research and Development Program of China 2018AAA0101400.

{\small
	\bibliographystyle{ieee_fullname}
	\bibliography{egbib}
}

\clearpage

\appendix

\renewcommand\thesection{\arabic{section}}

\noindent{\LARGE \textbf{Appendix}}
\vspace{5mm}

\section{More Details of Baselines}

In our main manuscript, we have briefly described several representative alignment-based methods, which we use as our baselines for validating the effectiveness of our MetaAlign. Here, we present more details of some baselines.

\noindent
\textbf{DANNPE.} As shown in Fig.~\ref{fig: supp_baselines}, \textbf{\textit{DANNPE}} differs from \textbf{\textit{DANN}} in two key aspects: 1) Similar to \cite{long2018cdane,cui2020gradually}, the predicted object classification probability/likelihood $C(G(\cdot))\in\mathbb{R}^K$ is treated as the input of domain discriminator $D$, instead of the output feature of $G(\cdot)$ in \emph{DANN}.
2) Following \cite{long2018cdane}, we prioritize the discriminator on those easy-to-transfer samples by re-weighting the samples based on the entropy of object class prediction, with the weight defined as $\omega(ent(\cdot))=e^{-ent(\cdot)}$, where  $ent(\cdot)$ denotes the entropy of the object class prediction.  As shown in Table 1 in our main manuscript, \textit{DANNPE} significantly outperforms \textit{DANN}.

\noindent
\textbf{MMD.} We directly add the MMD constraint \cite{borgwardt2006integratingmmd} on the output of $G(\cdot)$ to encourage the feature alignment between source domain and target domain data (see Fig.~2~(b) in our main manuscript). 
The complete MMD loss (\ieno, Eq.~(4) in the main manuscript) is formulated as:
	\begin{equation}
				\begin{split}	\mathcal{L}_{dom}=&\frac{1}{N_s}\sum_{i=1}^{N_s}\sum_{i'=1}^{N_s}\mathcal{K}(\mathbf{f}_i^s, \mathbf{f}_{i'}^s) + \frac{1}{N_t}\sum_{j=1}^{N_t}\sum_{j'=1}^{N_t}\mathcal{K}(\mathbf{f}_j^t, \mathbf{f}_{j'}^t) \\
				&- \frac{2}{N_sN_t}\sum_{i=1}^{N_s}\sum_{j=1}^{N_t}\mathcal{K}(\mathbf{f}_i^s, \mathbf{f}_{j}^t).
				\end{split}
	\label{l_mmd}
	\end{equation}
	where $\mathbf{f}_i = G(\mathbf{x}_i)$, and $\mathcal{K}(\mathbf{f}, \mathbf{f}')$ denotes a kernel function. 
	Following \cite{li2018domainMMDAAE}, we use the well-known characteristic kernel RBF, \ieno, $\mathcal{K}(\mathbf{f}, \mathbf{f}') = \exp{(-\frac{1}{2\sigma}\left\vert\left\vert \mathbf{f} - \mathbf{f}'\right\vert\right\vert^2)}$, where $\sigma$ is the bandwidth parameter \cite{li2018domainMMDAAE}.

\textbf{\begin{figure}[t]
	\centering
	\includegraphics[width=0.48\textwidth]{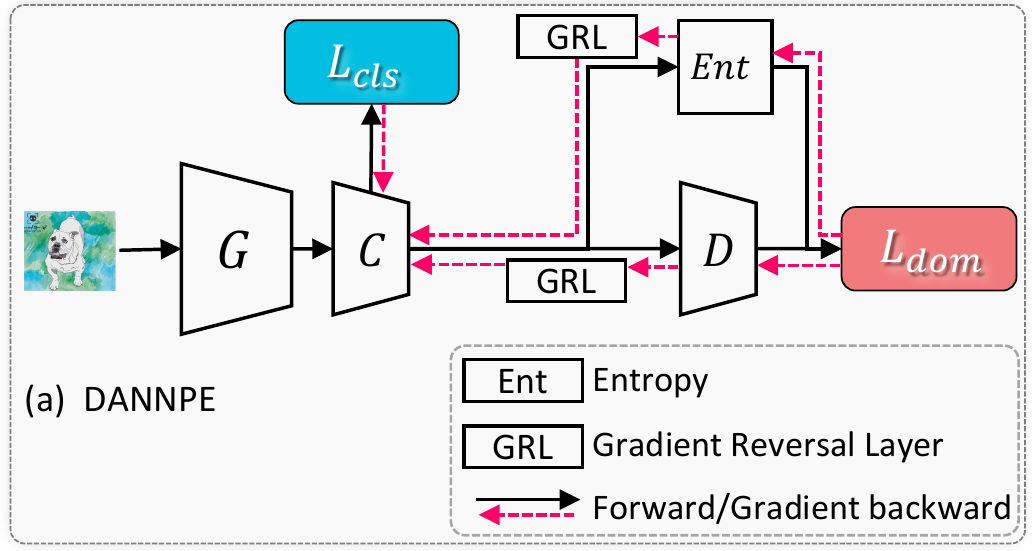}
	\vspace{-0.1cm}
	\caption{The pipeline of DANNPE.}
	\label{fig: supp_baselines}
\end{figure}
}

\textbf{\begin{figure*}[t]
	\centering
	\includegraphics[width=0.99\textwidth]{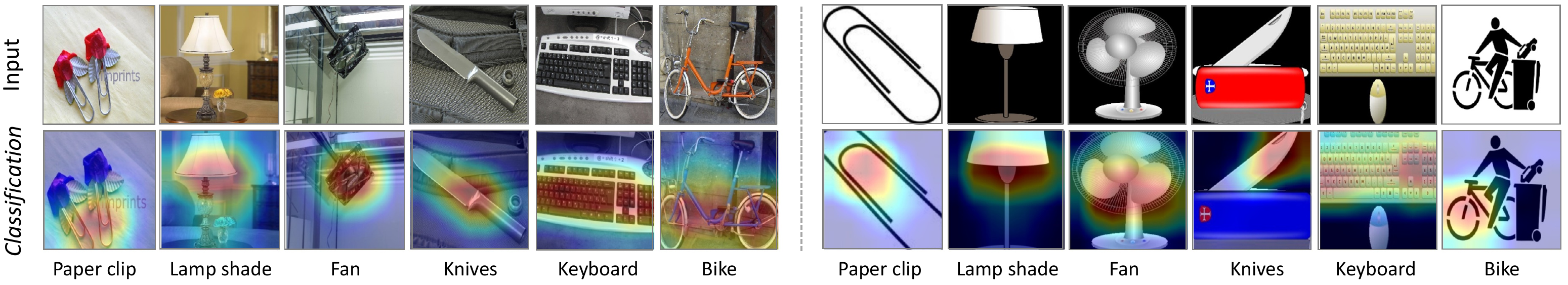}
	\vspace{-0.1cm}
	\caption{Visualization fo the Grad-CAMs \cite{selvaraju2017grad} w.r.t. the object classification task. The first row of left/right panels show the samples from source (Rw)/target (Cl) domains on Office-Home, while the second and third rows show the Grad-CAMs. The object classification task always focuses on foreground objects, which is also claimed in \cite{selvaraju2017grad, zhou2016cam}}
	\label{fig: grad_cams}
\end{figure*}
}

\begin{figure*}[t]
	\centering
\includegraphics[width=0.97\textwidth]{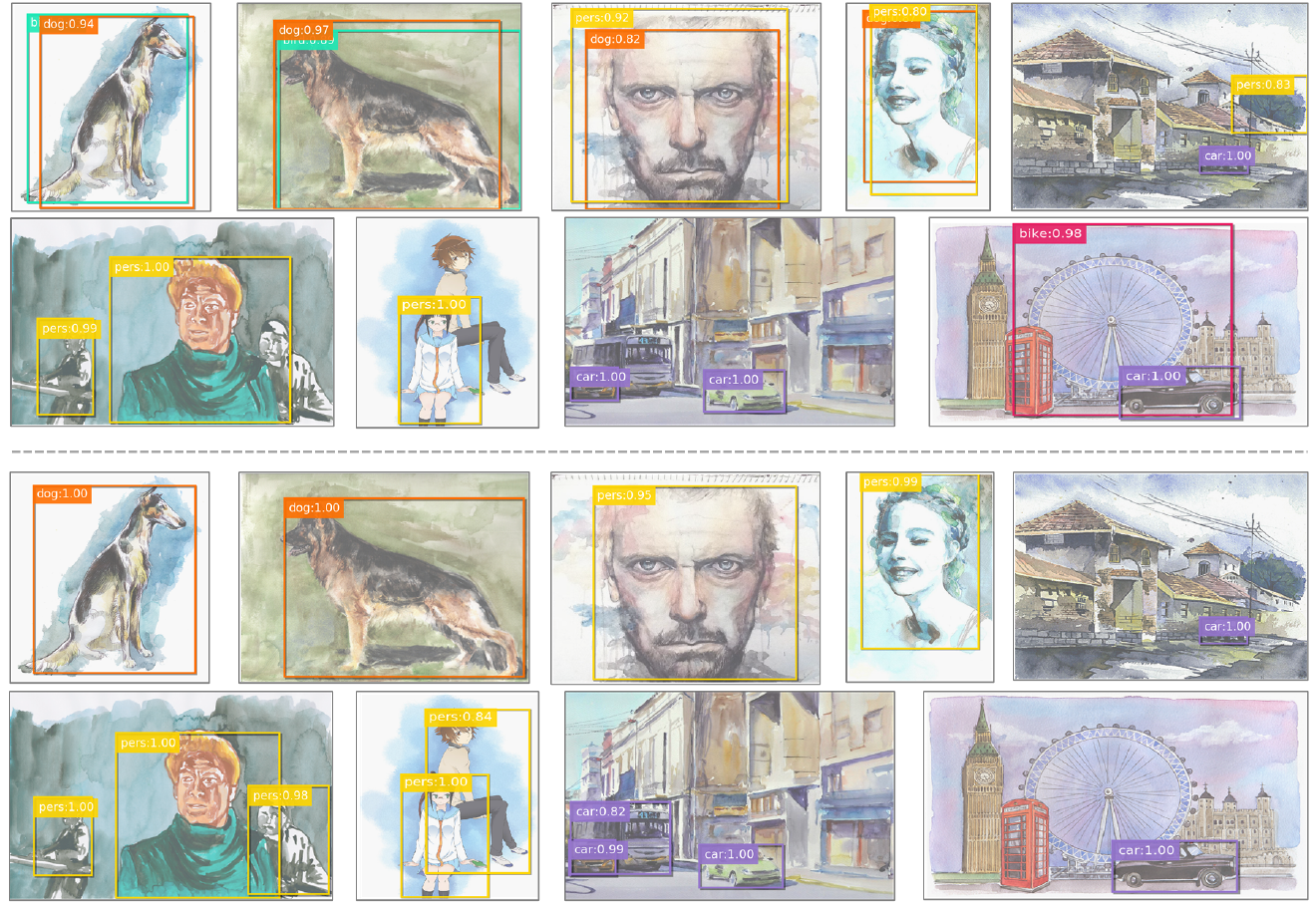}
	\vspace{-0.3cm}
	\caption{Object detection results on the target dataset Watercolor2k from (a) SW-DA (Baseline) (first two rows), and (b) SW-DA+MetaAlign (last two rows). }
	\label{fig: viz_detection_water}
\end{figure*}

For MMD-based UDA, similar to Eq.~(8) in the main manuscript, the optimization objective of MetaAlign is:
	\begin{equation}
		\begin{split}
			&\operatorname*{min}_{\theta, \phi_c, {\beta}} ~ \mathcal{L}_{dom}(\theta) \\ 
			&\phantom{arg}+ \mathcal{L}_{cls}\left(\left\{\theta_{m} - \alpha \beta_{m} \nabla_{\theta_m} \mathcal{L}_{dom}(\theta, \phi_d)\right\}_{m=1}^{M}, \phi_c\right)\\
			&\phantom{arg}+ \mathcal{L}_\beta(\beta).
		\end{split}
		\label{original_meta_weights_mmd}
	\end{equation}
	
\section{Experiments}

We describe more details on the implementation, datasets, settings, competitors, and present more experimental results.
    		
\subsection{UDA for Classification}

\noindent\textbf{Implementation Details.} We adopt ResNet-50 \cite{he2016deep} pre-trained on ImageNet \cite{krizhevsky2012imagenet} as the feature extractor for all baselines. Following \cite{cui2020gradually, long2018cdane}, the domain classifier/discriminator is composed of three fully connected layers with inserted dropout and ReLU layers for stable training, followed by a sigmoid function to output the domain classification result.
We divide the convolutional layers of the feature extractor $G$ into 4 groups (\ieno, $M=4$ in Eq.~(8)): the conv1 and conv2\_x as the first group, conv3\_x, conv4\_x, conv5\_x as the second to fourth groups respectively for simplicity. 

\noindent
\textbf{Grad-CAMs of Classification Task.} 
We illustrate the Grad-CAMs \cite{selvaraju2017grad} w.r.t. object classification task in Fig. \ref{fig: grad_cams}. As can be seen, the object classification task always focuses on the foreground objects, which is also validated in \cite{selvaraju2017grad, zhou2016cam}. 
    		
\subsection{UDA for Object Detection}

\noindent
\textbf{Datasets and Experimental Setting}. To simulate dissimilar domains, Pascal VOC \cite{pascalvoc2007, pascalvoc2012} and Watercolor2k \cite{inoue2018crosswatercolor} are treated as source and target domain respectively. 1) \textbf{Pascal VOC} \cite{pascalvoc2007, pascalvoc2012} is a well-known benchmark for object detection in real world scenario. In this dataset, 20 object classes with their corresponding bounding boxes are annotated. Following \cite{saito2019strongweak}, we employ the split setting which uses Pascal VOC 2007 and 2012 as training and validation. 2) \textbf{Watercolor2k} \cite{inoue2018crosswatercolor} is a collection of 2K watercolor images. It contains 6 categories in common with Pascal VOC. 1K images are used for training and the other 1K for testing.

As in previous works~\cite{chen2018domaindafaster, saito2019strongweak}, we set the shorter side of the image to 600 pixels following the implementation of Faster RCNN\cite{ren2015fasterrcnn} with ROI-alignment \cite{he2017maskrcnn}. The meta learning rate $\alpha$ is set to 0.01, which is 10 times the learning rate $\eta$.

\noindent
\textbf{Competitors.} We compare with the following methods: 1) \textbf{Source Only} trains model on source domain and directly tests on target domain. 2) \textbf{BDC-Faster} adopts the typical design of DANN, which takes the global features as input of the domain discriminator $D$ for adversarial learning. 3) \textbf{WST+BSR} \cite{kim2019selfwstbsr} constructs self-training on easy samples to reduce the negative effects of inaccurate pseudo-labels. 4) \textbf{MAF} \cite{he2019multimaf} incorporates multiple domain discriminators on hierarchical features. 5) \textbf{DT-UDA} \cite{inoue2018crosswatercolor} performs training on style-translated target images with predicted pseudo-labels. 6) \textbf{ATF} \cite{he2020domainatf} designs an asymmetric tri-way model to alleviate the collapse and out-of-control risk of the source domain. 7) \textbf{SW-DA} \cite{saito2019strongweak} aligns both global-level features and local-level features between the source and target domains by adversarial learning, which we take as our baseline for evaluating MetaAlign.

\noindent
\textbf{Visualization Results.} We have shown the performance comparison in Table 5 in our main manuscript. Here, we show the visualization of object detection results on the target dataset Watercolor2k \cite{inoue2018crosswatercolor} in Fig.~\ref{fig: viz_detection_water}. We can see that for the baseline scheme SW-DA, there are many false detections and missing detections. Thanks to the coordination between the domain alignment and the object detection optimization from our MetaAlign, the scheme SW-DA+MetaAlign achieves more accurate detections, where the false detections and missing detections are largely reduced.

\subsection{Domain Generalization}
\noindent
\textbf{Dataset and Settings.} \textbf{PACS} \cite{li2017deeperpacs} is a widely used benchmark for domain generalization. It contains 7 object categories from 4 domains (Photo, Art Painting, Cartoon and Sketch). We evaluate on this dataset under a commonly-used experimental protocol of leave-one-out \cite{li2017deeperpacs, carlucci2019domainjigen, li2019episodic}, where three domains are used for training and the remaining one is considered as the target domain. The domain discriminator $D$ of DANNPE here is kept the same as that for UDA classification, except that the final layer is a FC layer with 3 neurons instead of 1 for distinguishing the three source domains. 

\noindent
\textbf{Competitors.} 1) \textbf{\textit{AGG}} simply trains a model directly on the aggregation of all source domains. 2) \textbf{\textit{MMD-AAE}}\cite{li2018domainMMDAAE} equips an autoencoder with a MMD loss to train a domain-invariant encoder. 3) \textbf{\textit{CrossGrad}}\cite{shankar2018generalizingcrossgrad} is a typical data augmentation based DG method which perturbs in the input manifold to augment data. 4) \textbf{\textit{MetaReg}}\cite{balaji2018metareg}, 5) \textbf{\textit{MLDG}}\cite{li2017learningmldg} and 6) \textbf{\textit{MASF}}\cite{dou2019domainmasf} utilize meta-learning, which separate the samples into meta-train splits and meta-test splits, to mimic domain shift during training on source domains. 7) \textbf{\textit{JiGen}} imposes an auxiliary task of solving the Jigsaw puzzle on top of \textit{\textit{AGG}}. 8) \textbf{\textit{Epi-FCR}}\cite{li2019episodic} introduces a new episodic training strategy. 9) \textbf{\textit{MMLD}}\cite{matsuura2020domainmmld} predicts the pseudo domain labels and uses them for the adversarial domain learning.

\end{document}